%% file: main.tex
\documentclass[11pt,twocolumn]{article}

\usepackage[margin=0.75in]{geometry} % Slightly smaller margins for more content per page
\usepackage{multicol}
\usepackage[math]{alp}
\usepackage{graphicx}
\usepackage[numbers]{natbib}
\usepackage{hyperref}
\usepackage{color}
\usepackage{titlesec}

\titlespacing{\section}{0pt}{*2}{*1}
\titlespacing{\subsection}{0pt}{*1.5}{*0.8}
\title{Tools in the Loop: Quantifying Uncertainty of LLM Question Answering Systems That Use Tools}
\author{
  Panagiotis Lymperopoulos \\
  Tufts University \\
  \texttt{plympe01@tufts.edu}
  \and
  Vasanth Sarathy \\
  Tufts University \\
  \texttt{vasanth.sarathy@tufts.edu}
}
\date{}

\begin{document}

\twocolumn[
\maketitle
\begin{abstract}
Modern Large Language Models (LLMs) often require external tools, such as machine learning classifiers or knowledge retrieval systems, to provide accurate answers in domains where their pre-trained knowledge is insufficient. This integration of LLMs with external tools expands their utility but also introduces a critical challenge: determining the trustworthiness of responses generated by the combined system. In high-stakes applications, such as medical decision-making, it is essential to assess the uncertainty of both the LLM's generated text and the tool’s output to ensure the reliability of the final response. However, existing uncertainty quantification methods do not account for the tool-calling scenario, where both the LLM and external tool contribute to the overall system’s uncertainty. In this work, we present a novel framework for modeling tool-calling LLMs that quantifies uncertainty by jointly considering the predictive uncertainty of the LLM and the external tool. We extend previous methods for uncertainty quantification over token sequences to this setting and propose efficient approximations that make uncertainty computation practical for real-world applications. We evaluate our framework on two new synthetic QA datasets, derived from well-known machine learning datasets, which require tool-calling for accurate answers. Additionally, we apply our method to retrieval-augmented generation (RAG) systems and conduct a proof-of-concept experiment demonstrating the effectiveness of our uncertainty metrics in scenarios where external information retrieval is needed. Our results show that the framework is effective in enhancing trust in LLM-based systems, especially in cases where the LLM’s internal knowledge is insufficient and external tools are required.
\end{abstract}
\vspace{1em}
]

\input{sections/introduction}
\input{sections/related_works}
\input{sections/method}
\input{sections/experiments}
\input{sections/discussion}

\bibliographystyle{plainnat}
\bibliography{main.bbl}

\end{document}

%% file: sections/introduction.tex
\section{Introduction}
As large language models (LLMs) have been increasingly deployed in practical applications, their problem-solving capacity has enabled practitioners to combine them with external tools to augment their reasoning abilities and available information beyond their training distribution\cite{NEURIPS2023_d842425e}. This has greatly expanded LLMs' applicability beyond text-processing applications to problems involving complex decision making, which find applications in various critical domains such as medicine\cite{thirunavukarasu2023large} and the law\cite{lai2023large}. While recent work has developed metrics that help decide whether to trust LLM outputs \cite{kuhn2023semantic}, such metrics are not applicable to tool-calling systems directly, as they do not take into account the trustworthiness of the tool itself. This is especially important in highly specialized domains, where the LLM's own knowledge from training may not suffice and external tools are required.  

Uncertainty quantification has been a critical component in machine learning for determining when to trust model outputs. In particular, uncertainty measures such as predictive entropy, have proven useful in understanding whether a model's prediction is reliable, or whether the model is likely to produce unreliable predictions due to, for instance, distribution shift. Techniques such as Monte Carlo dropout \cite{gal2016dropout} and deep ensembles \cite{lakshminarayanan2017simple} have been applied to improve estimates of uncertainty in model predictions, especially in domains like healthcare, where model missteps can be costly. These methods allow practitioners to decide when further human intervention may be needed or when the system might fail. Extending this notion to LLMs, uncertainty quantification over token sequences or meanings has emerged as a tool for determining when to trust generated content \cite{kuhn2023semantic,farquhar2024detecting,malinin2020uncertainty}. Here, the uncertainty can be tied to the LLM’s internal representation of the sequence, which influences the model's capacity to produce coherent and accurate text.

Recently, generative AI agents have emerged as a powerful way to extend the capabilities of LLMs beyond text generation \cite{qin2304tool,liu2024toolace,NEURIPS2023_d842425e}, allowing these models to act as agents that interact with the world. By calling external APIs, models, or databases, LLMs can retrieve information, perform tasks, or even make decisions based on external computations. This paradigm shift has enabled LLMs to step into more agentic roles, executing actions that can impact real-world systems. However, this ability also comes with heightened safety and trust requirements\cite{hendrycks2021unsolved}. However, uncertainty quantification methods for agentic LLM systems are currently lagging behind, and advancing these techniques is crucial to making such systems safer for deployment in sensitive and high-stakes applications.

In this work, we take a first step towards increasing the trustworthiness of LLM agents by building a framework for quantifying the uncertainty in tool-calling question-answering (QA) systems. We begin by modeling the tool-calling LLM system and deriving uncertainty metrics for it, taking into account both the uncertainty of the LLM and the tool being called. In our work, we make the key assumption that we are operating in the white-box setting, where the uncertainty of external tools is known and we have access to model logits. To quantify uncertainty over the meaning of the combined system's answer, we adapt semantic entropy \cite{kuhn2023semantic}, into our framework. Finally, we make the uncertainty computation more efficient by deriving an approximation that makes our framework feasible in many practical settings. We also apply our framework to retrieval-augmented-generation (RAG) tasks. We evaluate our approach on two novel synthetic question-answering (QA) datasets requiring tool calling, derived from the well-known IRIS flower classification \cite{iris_53} and PIMA diabetes datasets. Finally, we also apply our framework to a simple retrieval-augmented generation \cite{gao2023retrieval,lewis2020retrieval} system and conduct a proof-of-concept experiment quantifying the uncertainty of answering questions from the BoolQ\cite{clark2019boolq} dataset of yes/no questions using an external database of references.

%% file: sections/related_works.tex
 \input{figures/system}
\section{Related works}
In the field of uncertainty quantification (UQ) for large language models (LLMs), a growing body of research has focused on developing techniques to estimate when these models might produce unreliable outputs. One approach involves using supervised models to predict uncertainty by leveraging the LLM’s internal states, such as logits or hidden activations, as well as ground truth labels from training data. For example, recent work\cite{liu2024uncertainty} has proposed training a separate uncertainty model based on LLM logits to improve the estimation of confidence in the generated outputs.

Another method involves modeling semantic uncertainty\cite{kuhn2023semantic,farquhar2024detecting}, which quantifies the LLM's uncertainty over the meanings of generated text rather than just over individual tokens. This approach is particularly useful for capturing more complex ambiguities. Semantic uncertainty has proven effective in detecting hallucinations and other inconsistencies in LLM-generated content, offering a pathway to more reliable responses. This method relies on entailment models to establish semantic classes of samples obtained from the LLM using the same prompt. Then, the uncertainty over semantic classes is estimated empirically using those samples.

While these uncertainty quantification techniques are crucial for understanding the reliability of LLMs, they do not directly account for the uncertainty introduced by external tools in tool-calling systems as they only involve the distribution over tokens (or meanings) expressing the system's final answer. Our framework for modeling tool-calling systems allows us to use these techniques for the LLM portion of the system, as well as combine them with the predictive uncertainty of the external tools in a principled way, thereby enabling more comprehensive uncertainty assessments in tool-augmented LLM systems.

Recent work on tool-calling LLMs \cite{NEURIPS2023_d842425e,qu2024tool} has largely focused on training large language models (LLMs) to invoke external tools in a structured and useful manner, as well as generating datasets that encourage models to learn how and when to use tools \cite{liu2024toolace}. Techniques such as fine-tuning LLMs on datasets designed for tool use, and building frameworks that allow LLMs to interact with APIs, databases, and external systems, have proven highly effective in enabling models to extend their capabilities beyond text generation. However, while these methods drive practical applications by enhancing the model’s functional scope, they do not address the issue of uncertainty quantification. 

In parallel, several tool-calling benchmarks have been introduced to evaluate the capabilities of LLM agents \cite{zhuang2023toolqa,peng2021revisiting}. These benchmarks test various aspects of agentic systems, including the ability to select appropriate tools when needed, call them with correct arguments, and utilize their results meaningfully. They often evaluate models on tasks that range from coding, API calls, and database queries to information retrieval and mathematical computations. 
While these benchmarks offer comprehensive assessments of an agent’s ability to use tools correctly, they do not focus on quantifying the uncertainty of the systems evaluated. In addition, many of the tools used in these benchmarks are often deterministic, or too complex so that assessing their uncertainty, especially in terms of predictive uncertainty, is challenging. As a result, we do not use them in our study and instead conduct controlled experiments aimed at assessing the efficacy of our framework in simple QA tasks that require calling tools. These more controlled environments allow us to study the more limited but still widely applicable setting when the tools provide known and reliable uncertainty estimates.  

%% file: figures/system.tex
\begin{figure*}
    \centering
    \includegraphics[width=1\linewidth]{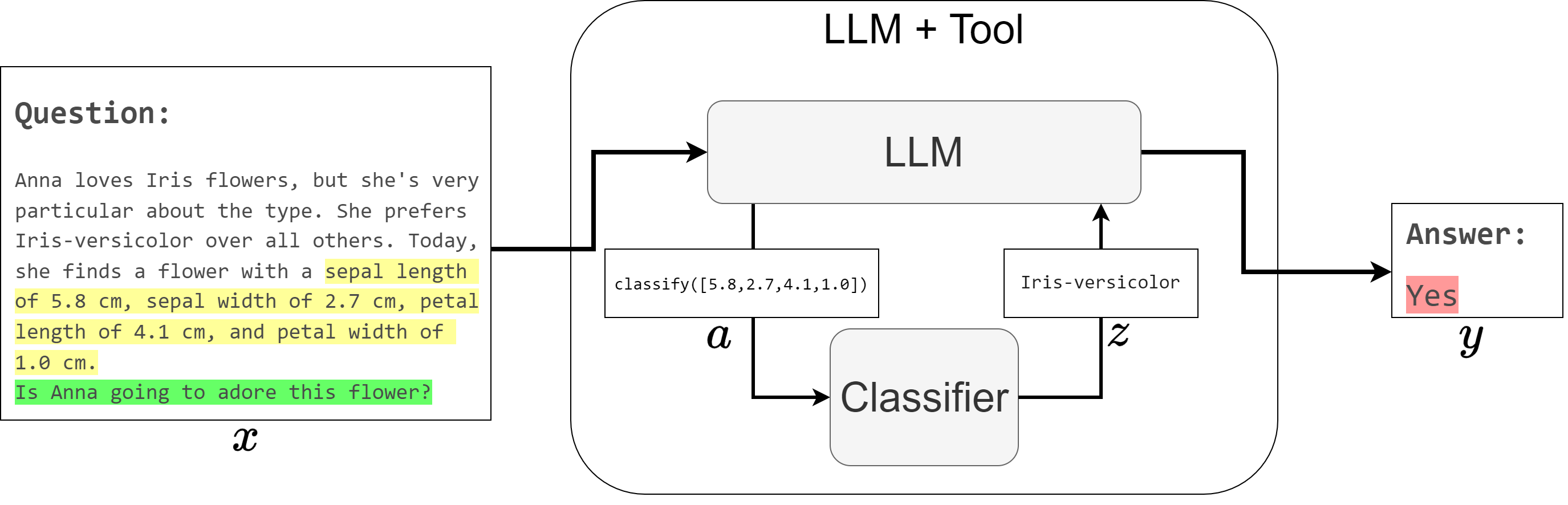}
    \caption{Illustration of our model of the LLM+tool system. The system receives an input prompt $x$, such as a question that requires a tool (e.g. a classifier) to answer. The LLM produces a tool call $a$ which acts as input to the tool. The tool produces output $z$, which in turn is mapped to a token sequence and provided to the LLM, alongside the original prompt. Finally, the LLM produces the final answer $y$. Yellow indicates the features that the LLM needs to extract for the tool call. Green indicates the final question the system needs to answer and red indicates the final answer provided by the combined system. Within this framework we can quantify the uncertainty of the final answer while taking into account the uncertainty of the classifier.}
    \label{fig:system}
\end{figure*}

%% file: sections/method.tex
\section{Method}
In this section we present our framework for uncertainty quantification in tool-using LLMs. Let $\calS$ be the set of all token sequences. Let $x \in \calS$ be a user-provided sequence representing the prompt to the LLM. Let $a\in\calA \subset \calS$ be a sequence representing an invocation of an external tool. Let $z\in\calZ$ be a result produced by the external tool, where $\calZ$ is the space of possible responses (e.g. $\calZ$ = \{0,1\} for binary classifiers). Finally, let $y\in\calS$ be a sequence corresponding to the response produced by the LLM after receiving the prompt and invoking the tool. Note that this formulation covers the case of multiple tools, as they can be encapsulated into a single tool with a more complex tool call $a$.

Our framework makes the following assumptions: 1) The final response $y$ is independent of the invocation of the tool $a$ given the tool response $z$. 2) The predictive entropy $H(z|a)$ of the tool $p(z|a)$ is known. 

% First, we derive the predictive entropy $H(y|x)$ of the distribution $p(y|x)$ represented by the combined LLM and tool system.
We model the tool-calling process as a sequential process encompassing two calls to the LLM and one call to the tool.
\begin{equation}
    p_\theta(y,z,a|x) = p_\theta(y|z,x)p(z|a)p_\theta(a|x),
    \label{eq:joint}
\end{equation}
Equation \eqref{eq:joint} shows the joint distribution over the variables in the system, where $\theta$ corresponds to the parameters of the LLM. Figure \ref{fig:system} illustrates our framework for modeling tool-calling LLM systems. First, the prompt $x$ is provided to the LLM to generate the tool call $a$. For instance, $x$ could contain the medical information of a patient and the description of two treatments for different patient subgroups. The LLM would first have to extract the appropriate features from the prompt and format them to produce $a$. Then $a$ is sent to the classifier $p(z|a)$ to classify the patient into appropriate treatment subgroups, represented by $z$. Finally, the LLM receives the tool output $z$ along with the original prompt and produces a final answer $y$. In practice, the classifier output may be deterministically mapped to a token sequence to make it more informative to the LLM. For instance, if the tool is a diagnostic model, $z$ may become "The patient is diagnosed with \textit{<disease>}". 

\subsection{Uncertainty quantification for tool-calling systems.} Within our framework, we quantify uncertainty using entropy. We now present a derivation for the predictive entropy $H(y|x)$ in our framework in terms of the known $H(z|a)$ and other terms.
\begin{align*}
    H(y|x) &= H(y,z,a|x) - H(z,a|x,y),\\
    %H(y|x) &= H(y|z,a,x) + H(z,a|x) - H(z,a|x,y),\\
    \begin{split}
     H(y|x) &= H(y|z,a,x)  + H(z|a,x) + H(a|x) \\&- H(z|x,y,a) - H(a|x,y).       
    \end{split}
\end{align*}
By the conditional independence in eq. \eqref{eq:joint} we obtain:
\begin{equation}
\begin{split}
    H(y|x) = &H(y|z,x)  + H(z|a) + H(a|x) \\&- H(z|y,a) - H(a|x,y).
    \end{split}
    \label{eq:pred_entropy}
\end{equation}

Equation \eqref{eq:pred_entropy} is the predictive entropy of the final answer produced by the combined LLM+tool system given the prompt. The first and third terms are predictive entropies over sequences of tokens produced by the LLM, which may be estimated empirically through samples. The second term is the predictive entropy of the tool, which is assumed to be known. The final two terms are based on the posteriors $p(z|y,a)$ and $p(a|y,x)$. These terms present a major challenge as they are very difficult to estimate: They involve marginalization over the space of token sequences, which is very large. Later in this section we discuss ways to tackle this intractability.

\paragraph{Semantic entropy for tool-calling systems.} 
\input{figures/ragsystem}
Semantic entropy \cite{kuhn2023semantic} is a technique for quantifying LLM's uncertainty over \textit{meanings} rather than token sequences. In summary, this is achieved by clustering sample responses from the LLM into a set of semantic classes $C$ and empirically estimating the distribution over meanings with samples. In our framework, we can apply the same principle in the final answer $y\sim p_\theta(y|x,z)$ produced by the system, after observing the tool result. Then, similarly to the predictive entropy case, we can derive semantic entropy within our framework: 
\begin{equation}
\begin{split}
        H(C|x) = &H(C|z,x)+ H(z|a)\\&+ H(a|x) - H(z|y,a) - H(a|x,y).
\end{split}
\label{eq:sem_entropy}
\end{equation}
As in the original formulation of semantic entropy, in our framework we can estimate $H(C|z,x)$ with samples. Note that we do not compute semantic entropy over the tool call $a$, as in our experiments, tools are simple to call for the LLMs. However, if there are many ways to express the same tool call (say, a question to a domain-specialized language model), then the entropy over equivalent classes of tool calls should be considered. 

\paragraph{Intractability of entropy calculation.}
Equations \eqref{eq:pred_entropy} and \eqref{eq:pred_entropy} express the predictive and semantic entropy within our framework for tool-calling LLMs. However, the negative terms $H(z|y,a)$ and $H(a|x,y)$ are problematic. Here we explain the challenge and discuss two approaches to deal with the intractability. $H(z|y,a)$ is the entropy of the posterior distribution $p(z|y,a)$, which expresses the probability of the classifier output, after the final answer of the LLM is observed. Similarly, $H(a|y,x)$ is the entropy of $p(a|y,x)$. Using Bayes' rule and marginalizing over $z$, we can obtain:
\begin{equation*}
   p(a | x, y) = \frac{\sum_{z \in \calZ } p_\theta(y | z, x) p(z | a) p_\theta(a | x)}{p_\theta(y |x) }.
\end{equation*}
Even though in some cases the space $\calZ$ is not necessarily large (e.g. when the tool is a binary classifier), the normalization term in the denominator is intractable as it requires marginalization of token sequences for the tool call $a$. As a result, we cannot compute it directly. Similar reasoning applies to $p(z|y,a)$. 

One approach to addressing the intractability is to estimate the posteriors empirically, by fitting a model to samples from the system. Consider the case where $p(z|a)$ is a binary classifier, which receives an input vector of $d$ features and outputs a binary label. We can fit a neural network $p_\phi(z|y,a)$ with parameters $\phi$ to tuples of $(y,z,a)\sim p_\theta(y,z,a|x)$, where inputs are vector embeddings of $z$ and $a$ and the output is a probability over $\calZ$. Similarly, we can use vector embeddings of $y$ and $x$ to predict continuous values for the features in $a$. However, empirically fitting the posterior distribution from a few samples can be difficult and the approximation not sufficiently accurate for quantifying uncertainty.

\paragraph{Estimating entropies} To estimate the entropy of the LLM's final answer $H(y|z,x)$, as well as the semantic entropy counterpart, we follow previous work and estimate them directly with samples. For a collection of samples of final answers $D = \{(y_i,z,x)\ | i=1 \ldots N\}$, the predictive entropy of the final answer is estimated by:
\begin{equation}
    H(y|z,x) \simeq \frac{1}{N} \sum_{i=1}^N \log p(y_i|z,x).
\end{equation}
Similarly, given a collection of samples $D' = \{(C_i,y_i,z,x)\ | i=1 \ldots N\}$, where $C_i\in C$ is the semantic class index of $y_i$, the semantic entropy of the final answer can be estimated by:
\begin{equation}
    H(C|z,x) \simeq \frac{1}{|C|} \sum_{j=1}^{|C|} \log \sum_{y\in C_j}p(y|z,x).
\end{equation}

\subsection{Strong Tool Approximation} By making some additional assumptions about the setting, we can obtain a much simpler approximation that is also more efficient to compute for equations \eqref{eq:pred_entropy} and \eqref{eq:sem_entropy}. We consider the \textit{strong tool setting} with the following conditions:
\begin{itemize}
    \item There is strong dependence between $y$ and $z$. This means that the question is very difficult for the LLM to answer without knowing the tool output, but almost obvious given the tool output. At the same time, it is relatively easy to infer $z$ if $y$ is known.
    \item All of the information about $a$ is contained in $x$. In other words, since it is possible to determine $a$ from $x$, knowledge of $y$ is unlikely to provide much benefit in predicting $a$. 
\end{itemize}

Note that these conditions do not express a requirement on the quality of the tool, but is rather a feature of the application domain. We consider this a realistic setting as it is in-line with expected uses of tool-calling systems: if a tool is not necessary for the application, and if the information for calling it is not to be found in the input, it is unlikely to be implemented in the first place. For example, in a medical application that suggests treatments to medical professionals, the external tool could be a complex diagnostics model over a patient's blood-work. The LLM may use its general knowledge to suggest treatments after knowing the diagnosis, but is unlikely to provide an accurate diagnosis without the use of an expert tool. At the same time, knowing a patient's treatment plan is not useful in formulating the call to the diagnostic model, given the patient information. 

\input{figures/data}
Examining equations \eqref{eq:pred_entropy},\eqref{eq:sem_entropy} in this setting allows us to make some simplifications. First, we notice that since $y$ strongly depends on $z$, the distribution $p(z|y,a)$ is going to be low-entropy, contributing only slightly to the computation. Additionally, since $y$ is not very informative about $a$, the values $H(a|x)$ and $H(a|x,y)$ are likely to be similar and cancel out. We therefore arrive at the Strong Tool Approximation (STA) of Predictive and Semantic Entropy ($STA_P$, $STA_S$):
\begin{align}
    STA_P(x) &= H(y|z,x)+ H(z|a),\\
    STA_S(x) &= H(C|z,x)+ H(z|a).
\end{align}
These metrics are simple to compute, amounting to only additively combining the entropy of the LLM's final answer, which can be estimated using existing methods, and the entropy of the tool response, which is assumed to be known. While the STA ignores some terms, notably the entropy of the tool call, in our experiments we show that in the strong tool condition, these metrics can be more reliable than fitting the posterior terms with a few samples, which is error-prone. 

\paragraph{Application to RAG} With some modification, we can apply our framework to QA with RAG, where the system first retrieves documents relevant to a query from a corpus and then answers the query with the relevant documents inserted to the context. In this setting there is no tool call, since the document retrieval happens automatically. 
\begin{equation}
    p_\theta(y,z|x) = p_\theta(y|z,x)p(z|x),
    \label{eq:joint_rag}
\end{equation}

We consider a soft form of RAG, where the retrieval system defines a categorical distribution $p(z|x)$ over documents in the corpus. Here, the space of $z$, $\calZ$ is the corpus of documents. There are many ways to define $p(z|x)$, for instance, vector embeddings can be computed for the query and documents in the corpus, and cosine similarity can be used to compute logits over the documents, as is common in RAG applications. Then samples can be drawn from this distribution to retrieve relevant documents. 

The predictive entropy for the system is:
\begin{equation}
    H(y | x) = H(y | z, x) + H(z | x) - H(z | y, x).
    \label{eq:rag_entropy}
\end{equation}
As previously, if the dependency between $z$ and $y$ is strong, we can compute the STA metrics, using a similar formulation for the semantic entropy. However, estimating $H(z | y, x)$ empirically may be very difficult, especially since the document corpus may be large.  

In general for most queries, only a small number of documents are likely to be relevant, making the distribution low-entropy. If no relevant documents are available, then most documents will have similar probabilities, making the retrieval distribution high-entropy. Since the LLM is more likely to answer correctly if a relevant passage is found, the tool entropy is likely to be helpful in predicting the correctness of the overall system response.

% \begin{equation}
%     H(C|x) \simeq \frac{1}{|C|}\sum_{i=1}^{|C|}\log \sum_{y\in C_i}p(y|z,x)+ H(z|a)
% \end{equation}

%% file: figures/ragsystem.tex
\begin{figure*}
    \centering
    \includegraphics[width=1\linewidth]{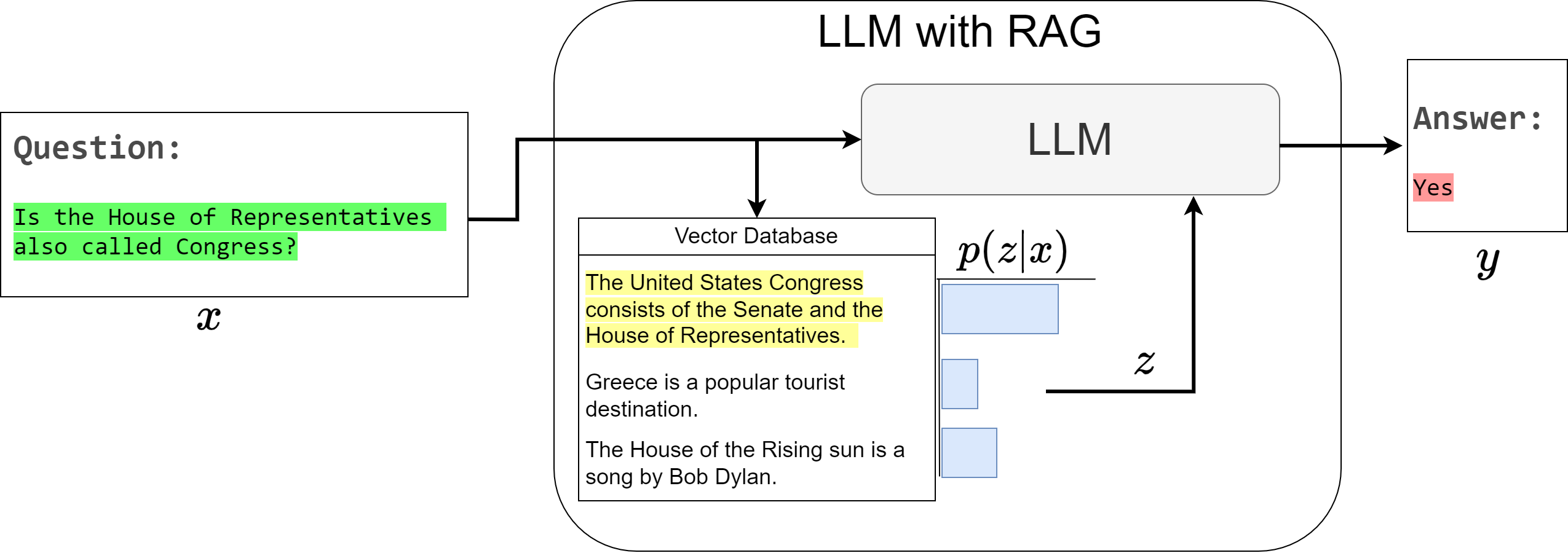}
    \caption{Illustration of our framework applied to RAG. The system receives an input question $x$, shown in green, that requires additional documents to answer. In our framework, we describe the document retriever as a categorical distribution over documents. The system samples relevant documents $z$ from that distribution  which are added to the LLM context. Finally, the LLM produces the final answer $y$. Yellow indicates the most relevant passage. The bar plot shows that in this example the retrieval distribution is low entropy, so uncertainty in the retrieval is low. Red indicates the final answer provided by the combined system. Within this framework we can quantify the uncertainty of the overall system answer, taking into account the uncertainty of the retreival system.}
    \label{fig:system}
\end{figure*}

%% file: figures/data.tex
\begin{figure*}
    \centering
    \includegraphics[width=0.7\linewidth]{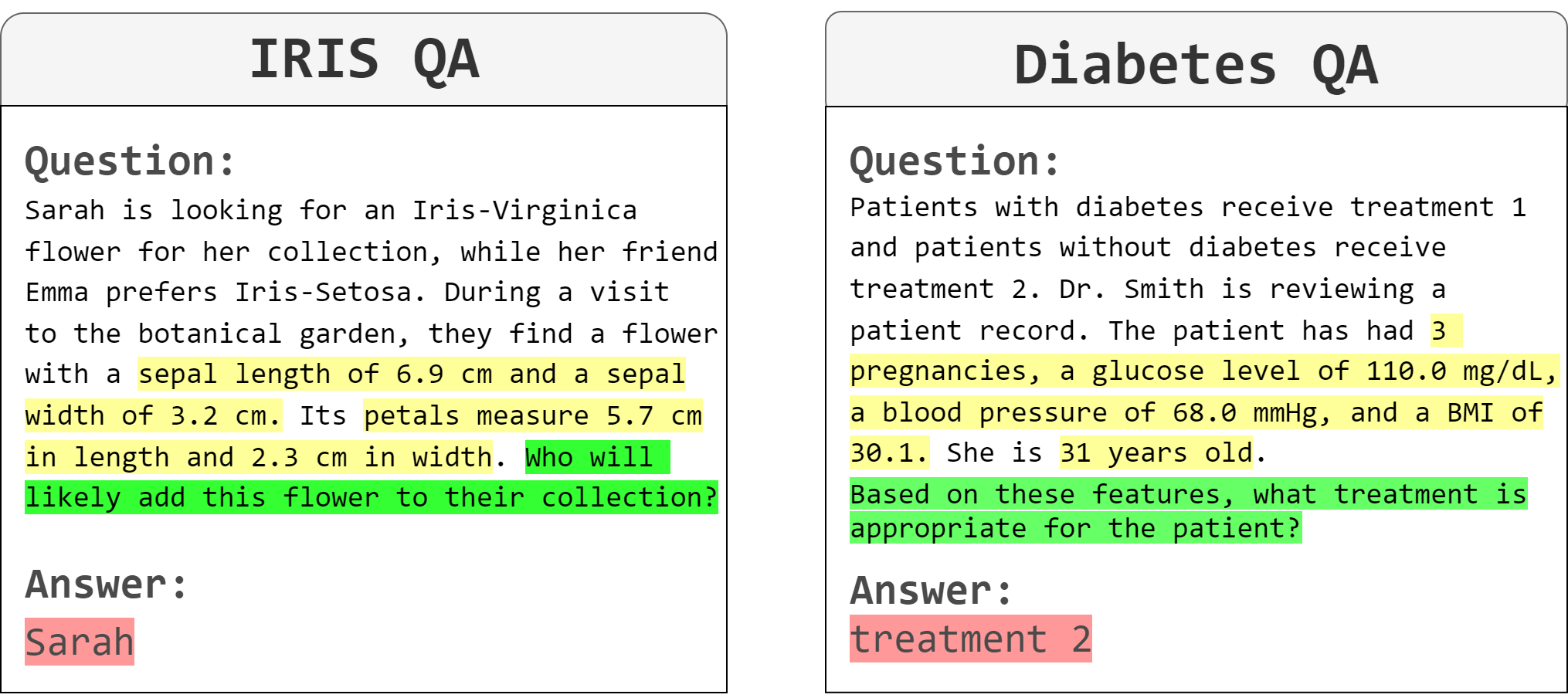}
    \caption{Samples from the IRIS QA and the Diabetes QA datasets. Yellow indicates the portion of the question that the LLM needs to extract features for the tool call from. Green indicates the question that the system needs to answer based on the prompt and the tool response. Red indicates the answer.}
    \label{fig:data}
\end{figure*}

%% file: sections/experiments.tex
\input{tables/iris}
\input{tables/pima}

\section{Experiments}
In this section we present experiments validating our framework and the derived metrics. First, we discuss the creation of the synthetic datasets we use for evaluating the metric in the QA with tool calling setting. Then we describe and discuss our experiments on those datasets and the RAG QA task.

\subsection{Datasets}
As discussed earlier, current datasets for evaluating tool calling LLM agents are not well suited for the study of uncertainty quantification, due to the complexity of the queries and tools (e.g. writing code) which are aimed at evaluating the ability of systems to take advantage of tools to answer queries, rather than to quantify uncertainty. As a result we create 2 novel synthetic QA datasets based on the IRIS\cite{iris_53} and Pima Indians diabetes\cite{smith1988using} classification datasets.

We create a QA dataset from each machine learning dataset as follows: First, we fit a gaussian mixture model to each class and sample synthetic datapoints to ensure that the LLMs have not seen these data points during their training. Then, we sample 150 points, evenly split among each class. We preprocess the data to better match the formatting of the original dataset (e.g. rounding to 1 decimal place). We construct a few example questions by hand, where we pose a natural language question related to the classification result. We also ensure to include all features in the question and that the answer strongly depends on the classification result. Finally, we make sure that the question can be answered briefly, typically in one word. Using the constructed examples we query GPT-4o with few shot prompting to construct new questions based on our sampled points and examine the results manually to ensure quality. We include the exact prompt for each dataset in the supplemental material alongside our code submission. As a post-processing step, we "anonymize" the dataset, replacing class names with identifiers (e.g. Iris-setosa with flower\_type\_1) to encourage the LLM's to base their response on the tools rather than their general knowledge. We are releasing our datasets alongside the code to generate them as part of our contribution with this work. 

For the RAG QA task, we sample 150 questions randomly from the BoolQ dataset \cite{clark2019boolq} along with their corresponding wikipedia excerpts containing their answers. We then mix the passages for half the questions with a corpus of other excerpts from wikipedia collected from an existing wikipedia QA dataset\cite{smith2008question} to create the document corpus for retrieval.

\paragraph{Evaluation Methodology}
To evaluate our framework we follow prior work and use AUROC as our evaluation metric \cite{kuhn2023semantic,band2022benchmarking}. The intuition behind this criterion is to treat the uncertainty metric as a score expressing whether we should trust the LLM output, with high entropy values indicating we should not, and low entropy values indicating that we should. Then, we can compare this score against the answer accuracy of the system. In short, if we trust the system when it answers correctly, that corresponds to a true positive prediction. If we trust it when it is wrong, this corresponds to a false positive prediction. AUROC enables us to evaluate the quality of metrics across decision thresholds.

In our experiments we compute the following uncertainty metrics derived from our framework:
\begin{itemize}
    \item $STA_P$: The strong tool approximation applied to predictive entropy.
    \item $STA_S$: The strong tool approximation applied to semantic entropy.
    \item Sem. Entropy: The semantic entropy in our framework (eq \eqref{eq:sem_entropy}), computed with empirically estimated posteriors.
    \item Pred. Entropy: : The predictive entropy in our framework (eq \eqref{eq:pred_entropy}), computed with empirically estimated posteriors.
\end{itemize}
We also compare these metrics against naive baselines that do not combine sources of uncertainty:
\begin{itemize}
    \item  Sem. Entropy FA: The semantic entropy of the final answer produced by the LLM, given the tool output.
    \item Pred. Entropy FA: The  predictive entropy of the system's final answer, given the tool output. 
    \item Tool Entropy: The entropy of the tool itself.
\end{itemize}

We run our experiments on a server with one NVIDIA A600 GPU, an AMD Ryzen Threadripper PRO 5955WX CPU with 16-Cores and 130GB of RAM. Due to computational limitations, we only experiment with smaller LLMs. However, the framework is applicable as is to larger LLMs and prior work suggests results are likely to generalize to larger models as well \cite{kuhn2023semantic}. We use the instruction-tuned variants of the models, as they are better suited for the tool calling setting.

\subsection{Experiment 1: Quantifying uncertainty in tool-calling QA}
In this experiment we evaluate our framework in quantifying the uncertainty of QA tool-calling LLM systems using the IRIS QA and Diabetes QA datasets. 
\paragraph{Experiment Settings} To compute uncertainties we take 10 samples of the final answer over 3 runs of the combined system to estimate uncertainties. We use the samples from the first 30 examples from each dataset to train the posterior models to estimate semantic and predictive entropy. Additional training details for the posterior estimates are available in the supplemental material.  We evaluate all metrics on the remaining 120 examples. We prompt using few-shot prompting with 3 examples both for the tool-calling step and for the final answer. Example prompts are available in the supplemental material alongside our code submission. The classifier called by the LLM is a simple lookup table with added uncertainty. For half the samples, the classifier uniformly samples a class to output. For the other half, it gives high probability to the correct class and low probability to the rest. 
\input{tables/rag}
\paragraph{Results} Table \ref{tab:results_iris} shows experimental results in the IRIS QA dataset for three models. Overall in this dataset the metrics based on the STA seem to perform better than the alternatives. This indicates that the conditions identified for STA to apply are satisfied, and that the noise introduced by the approximate posteriors is reducing the quality of the computed metrics. The baseline approaches computing entropies only over the final answer are in most cases much less informative, as they only consider uncertainty from one part of the system. Similarly, the tool entropy is also not as informative as the STA metrics, since they combine uncertainty from the whole system. 

Table \ref{tab:results_pima} shows experiment results from the Diabetes QA dataset. Overall, the results are consistent with those from the IRIS QA dataset: $STA_S$ and $STA_P$ outperform the other metrics in almost every case and the empirical approximation of semantic and predictive entropies seems to be too noisy. 

There is some surprising variation, particularly with the final answer entropies occasionally being more informative than expected. This variation may be due to how different models make use of the tool result, which may also depends on prompt engineering. If a model is prone to ignore the tool result for some samples, the entropy of the final answer may be more informative of the response accuracy than if the model is very faithful to the tool. Still, given the same prompts, $STA_S$ significantly outperforms the other metrics in both datasets, showing that when the STA assumptions hold, it is a viable and easy to compute metric for predicting response accuracy,

\subsection{Experiment 2: Proof of concept application to RAG QA}
In this experiment, we apply our framework to the RAG QA setting and quantify the uncertainty of the combined retrieval and LLM system. 
\paragraph{Experiment settings} To compute uncertainties, we take 10 samples of the final answer from the LLMs after retrieving documents. We construct the document retrieval system in two stages: First, we compute the cosine similarity of embeddings for the input question with the entire corpus of documents and select the top K items. Then, we use the scores over those K items as logits of a categorical distribution, and sample M documents from that distribution. In our experiments we use K=5 and M=1. We use the entropy of the categorical distribution as the entropy of $p(z|x)$. Additional implementation details are available in the supplemental material. 

In this experiment we only compute the STA and the baseline metrics, as computing the posterior term for the retrieval system is much more complex than for a classifier model, and obtaining a good approximation is unlikely.  
\paragraph{Results}
Table \ref{tab:results_rag} shows the experiment results for the RAG QA task using questions from the BoolQ dataset. For both Llama models, the $STA_S$ metric outperforms $STA_P$ and the baseline metrics, providing a more informative signal for when to trust the retrieval+LLM system. However, the performance difference is not as impressive as in the IRIS QA and Diabetes QA datasets, and in the case of Mistral-7B-Instruct model, there is no improvement, perhaps even a small degradation. This may be because in the RAG QA problem, the conditions of the \textit{strong tool setting} are less applicable than in the previous task. In our experiment, this is specifically the case with the first condition,  since there is no tool call. While the retrieval component provides useful information to the LLM, information without which answering the question is more difficult, making use of the retrieved information requires additional reasoning. As a result, the dependence between the final answer $y$ and document retrieved $z$ is not as strong, making the approximation worse. 

Another reason for the smaller improvement of the STA metrics over the baseline metric may be that, since the questions in this dataset are more general in nature than the specialized questions of previous tasks, the LLM has some general knowledge that it can use to answer the questions, or may have seen some of these questions during training. As a result, whether or not the retrieval system found the right document has a smaller influence on the final answer produced. Still, since the LLM does base its responses on the documents in some occasions, tool entropy is still somewhat predictive of the response accuracy and combining the metrics through the STA tends to improve our ability to predict response accuracy. 

%% file: tables/iris.tex
\begin{table*}[ht]
\tiny
\centering
\renewcommand{\arraystretch}{1.5}  % Adjust row spacing
\setlength{\tabcolsep}{3pt}       % Adjust column spacing
\begin{tabular}{|l|c|c|c|c|c|c|c|}
\hline
\textbf{Model} & $\mathbf{STA_S}$ & $\mathbf{STA_P}$ & \textbf{Sem. Entropy} & \textbf{Pred. Entropy} & \textbf{Sem. Entropy FA} & \textbf{Pred. Entropy FA} & \textbf{Tool Entropy} \\
\hline
Meta-Llama-3-8B-Instruct & 0.845 & 0.824 & 0.766 & 0.710 & 0.615 & 0.553 & 0.806 \\
Meta-Llama-3.1-8B-Instruct & 0.752 & 0.668 & 0.640 & 0.603 & 0.642 & 0.529 & 0.692 \\
Mistral-7B-Instruct & 0.786 & 0.718 & 0.712 & 0.683 & 0.667 & 0.605 & 0.753 \\
\hline
\end{tabular}
\caption{Experimental results from the IRIS QA dataset. Columns show AUROC of uncertainty metrics for different models. High values indicate the metric is predictive of response accuracy. In this dataset, the metrics based on the strong tool approximation outperform the other metrics. Interestingly, semantic and predictive entropy computed via empirical posteriors are not as predictive of accuracy, likely due to poor approximation. The entropy of the tool itself is informative, but the STA metrics offer additional benefits.}
\label{tab:results_iris}
\end{table*}

%% file: tables/pima.tex
\begin{table*}[ht]
\small
\centering
\renewcommand{\arraystretch}{1.5}  % Adjust row spacing
\setlength{\tabcolsep}{3pt}       % Adjust column spacing
\begin{tabular}{|l|c|c|c|c|c|c|c|}
\hline
\textbf{Model} & $\mathbf{STA_S}$ & $\mathbf{STA_P}$ & \textbf{Sem. Ent.} & \textbf{Pred. Ent.} & \textbf{Sem. Ent. FA} & \textbf{Pred. Ent. FA} & \textbf{Tool Ent.} \\
\hline
Meta-Llama-3-8B-Instruct & 0.791 & 0.730 & 0.635 & 0.652 & 0.663 & 0.634 & 0.642 \\
Meta-Llama-3.1-8B-Instruct & 0.675 & 0.662 & 0.656 & 0.674 & 0.515 & 0.580 & 0.664 \\
Mistral-7B-Instruct & 0.782 & 0.702 & 0.671 & 0.666 & 0.516 & 0.570 & 0.781 \\
\hline
\end{tabular}
\caption{Experimental results from the Diabetes QA dataset. Columns show AUROC of uncertainty metrics for different models. High values indicate the metric is predictive of response accuracy. Similarly to the IRIS QA results, the metrics based on the strong tool approximation tend to outperform the other metrics in most cases. Also here, likely due to poor approximations, the metrics using empirical estimates do not perform as well as those based on the STA as they are more noisy.}
\label{tab:results_pima}
\end{table*}

%% file: tables/rag.tex
\begin{table*}[ht]
\small
\centering
\renewcommand{\arraystretch}{1.5}  % Adjust row spacing
\setlength{\tabcolsep}{10pt}       % Adjust column spacing
\begin{tabular}{|l|c|c|c|c|c|}
\hline
\textbf{Model} & $\mathbf{STA_S}$ & $\mathbf{STA_P}$ & \textbf{Sem. Ent. FA} & \textbf{Pred. Ent. FA} & \textbf{Tool Entropy} \\
\hline
Llama-3.1-8B-Instruct & 0.675 & 0.646 & 0.662 & 0.622 & 0.635 \\
Meta-Llama-3-8B-Instruct & 0.648 & 0.645 & 0.570 & 0.575 & 0.629 \\
Mistral-7B-Instruct & 0.668 & 0.705 & 0.502 & 0.711 & 0.667 \\
\hline
\end{tabular}
\caption{Experimental results in the RAG QA task. In this setting, the STA metrics still offer a benefit by combining sources of uncertainty, though the improvement is smaller than in other tasks. Nevertheless, they outperform baseline approaches in almost every case. One possible explanation for the smaller improvement, is that in the RAG setting, the STA is not as applicable as in other settings. This is because the retrieved document may include useful information for answering the question, but significant additional reasoning may be necessary to obtain a correct answer. }
\label{tab:results_rag}
\end{table*}

%% file: sections/discussion.tex
\section{Discussion}

Our proposed framework for uncertainty quantification in tool-calling LLM systems presents a flexible and extensible approach to modeling uncertainty. One of its key strengths is that it is agnostic to the specific techniques used to compute entropies over token sequences or meanings. This flexibility allows for the integration of future advancements in uncertainty quantification without requiring significant changes to the underlying framework. For instance, emerging methods for better capturing LLM uncertainties, such as refined logit-based entropy estimation or latent semantic representation models, can be seamlessly incorporated into the framework as drop-in replacements. This modularity ensures that uncertainty quantification in tool-calling LLM systems continues to track developments in the broader field of LLM uncertainty estimation, and offers the potential for improving predictive accuracy by leveraging these innovations.

Our framework is also compatible with various architectures of tool-calling systems, making it suitable for a wide range of applications. For instance, in our experiments we consider the case where a tool is called for every query to the LLM. However, in practice, we may want to let the LLM decide whether to call a tool. This can be easily accommodated by adding a special entry in the tool call $a$, representing a null tool with zero entropy. In cases where the LLM has enough knowledge to answer a query without external assistance, this entry allows the system to omit the tool call while keeping the entropy calculation consistent.

Moreover, the framework can handle more complex architectures, such as those involving multiple tool calls. When more than one tool may be called, the decision to call the correct tool can be incorporated into the action space $a$, allowing the LLM to choose between tools dynamically. This ensures that our framework scales to scenarios where multiple tools are available, each with different competencies, and where tool selection plays a critical role in generating accurate answers. One challenge to consider is that depending on the application and tools, the STA may be weaker in this setting. This is because it may be harder to decide how to call the tool given the input, making the gap between $H(a|x)$ and $H(a|x,y)$ larger. In that case, more sophisticated methods for estimating the posteriors may be required, and the full metrics used.

Additionally, our framework supports parallel tool calls, in which the output space $\mathcal{Z}$ is formed by combining the outputs of all possible tools. This generalization allows the LLM to process information from multiple tools simultaneously, further enhancing its capabilities while still fitting within the entropy-based uncertainty quantification system. However, as with the previous case, this may make the STA weaker. 

Another possible extension to the framework is considering multiple rounds of tool-calling, interleaved with text generation for reasoning. In such cases, our framework can be extended by adding additional intermediate variables, with dependencies analogous to those of $x,z$ and $y$. However, it is worth noting that the strong tool approximation (STA) is also likely to degrade in that setting, as errors introduced by early rounds of tool calls accumulate, potentially leading to a loss in predictive power. Addressing this degradation may require further refinement of the approximation techniques, especially in domains requiring long chains of tool calls.

Turning to our experimental results, the application of our framework to retrieval-augmented generation (RAG) systems yielded smaller performance gains compared to the tool-calling QA experiments. We hypothesize that this is because the LLMs used in the RAG task were able to answer some of the questions directly from their internal knowledge, without heavily relying on the retrieved documents. In contrast, tool-calling settings typically involve specialized domains where the LLM's knowledge is insufficient, and uncertainty metrics derived from our framework offer a clearer benefit in predicting response accuracy. We expect that in more domain-specific RAG tasks, where the LLM’s internal knowledge is less complete, the STA-based uncertainty metrics will demonstrate a greater advantage.

One limitation of our methodology was in computing the full semantic and predictive entropy metrics as derived in our framework. While we attempted to empirically estimate the posterior distribution, the small datasets and the inherent difficulty of the task meant we had limited success. We still consider their inclusion necessary however as advancements in inference with LLMs may make computing such terms more feasible in the future.

Lastly, we acknowledge that our experiments were conducted using smaller LLMs due to computational resource limitations. Despite this, there is nothing inherent in our framework that prevents its application to larger models. Future work should explore applying our framework to larger LLMs to confirm its scalability and evaluate its performance in more demanding scenarios.

\paragraph{Broader Impact Statement} This work paves the way towards more responsible and trustworthy use of tool-calling LLM systems by providing a framework for quantifying uncertainty, which can help mitigate user overtrust in these systems and encourage more responsible use. By identifying when a system’s predictions are likely unreliable, we can reduce the risk of over-reliance on AI in critical applications, such as healthcare or legal decision-making, ultimately promoting safer AI deployment. However, the computational cost of computing uncertainty metrics, even approximate ones, is a significant drawback. Current methods require taking multiple samples from the system, which increases the already high energy demands of LLMs and raises concerns about the environmental impact of such models. As LLMs become more widely used, balancing the benefits of uncertainty quantification with the environmental costs will be an important challenge for the field.
\section{Conclusion}

In this work, we present a novel framework for uncertainty quantification in tool-calling large language models (LLMs). 
A key aspect of our approach is its flexibility, allowing for a variety of entropy computation methods to be integrated into the framework. This allows us to apply the semantic uncertainty idea to the tool-calling setting and derive efficient approximations, enabling practical deployment of uncertainty quantification in real-world scenarios. 

We validate our framework on two new synthetic QA datasets, derived from two well-known machine learning datasets, which demonstrate the framework’s capability to handle tool-calling settings. Additionally, we conduct a proof-of-concept experiment in a retrieval-augmented generation (RAG) scenario, showing that while the improvement from our uncertainty metrics was smaller in the RAG setting, the framework holds promise for specialized domains where LLM knowledge is limited. Finally, we provide extensive discussion of how our framework relates to different LLM agent architectures for tool-calling, how it can be extended to accommodate them and possible challenges in that direction.